\documentclass[10pt,twocolumn,letterpaper]{article}

\usepackage{algorithm}
\usepackage{algorithmic}
\usepackage{amssymb}
\usepackage{multirow}
\usepackage{subfigure}
\usepackage{amsmath}

\usepackage[pagenumbers]{cvpr} 

\usepackage[dvipsnames]{xcolor}

\definecolor{cvprblue}{rgb}{0.21,0.49,0.74}
\usepackage[pagebackref,breaklinks,colorlinks,citecolor=cvprblue]{hyperref}

\def\paperID{*****} 
\def\confName{CVPR}
\def\confYear{2024}

\title{OD-VAE: An Omni-dimensional Video Compressor \\
for Improving Latent Video Diffusion Model}

\author{Liuhan Chen\textsuperscript{\rm 1, \rm 3, *}, Zongjian Li\textsuperscript{\rm 1, \rm 3, *}, Bin Lin\textsuperscript{\rm 1,\rm 3}, Bin Zhu\textsuperscript{\rm 1,\rm 3}, Qian Wang\textsuperscript{\rm 1,\rm 3}, 
\\
Shenghai Yuan\textsuperscript{\rm 1,\rm 3}, Xing Zhou\textsuperscript{\rm 3}, Xinhua Cheng\textsuperscript{\rm 1,\rm 3}, Li Yuan\textsuperscript{\rm 1, \rm 2, \dag}  \\
\\
\textsuperscript{\rm 1}Peking University,  \textsuperscript{\rm 2}Peng Cheng Laboratory, \textsuperscript{\rm 3}Rabbitpre Intelligence \\
\\
{\tt\small liuhanchen@stu.pku.edu.cn, yuanli-ece@pku.edu.cn}
}

\begin{document}
\maketitle
\begin{abstract}
Variational Autoencoder (VAE), compressing videos into latent representations, is a crucial preceding component of Latent Video Diffusion Models (LVDMs).
With the same reconstruction quality, the more sufficient the VAE's compression for videos is, the more efficient the LVDMs are.
However, most LVDMs utilize 2D image VAE, whose compression for videos is only in the spatial dimension and often ignored in the temporal dimension.
How to conduct temporal compression for videos in a VAE to obtain more concise latent representations while promising accurate reconstruction is seldom explored.
To fill this gap, we propose an omni-dimension compression VAE, named OD-VAE, which can temporally and spatially compress videos.
Although OD-VAE's more sufficient compression brings a great challenge to video reconstruction, it can still achieve high reconstructed accuracy by our fine design.
To obtain a better trade-off between video reconstruction quality and compression speed, four variants of OD-VAE are introduced and analyzed.
In addition, a novel tail initialization is designed to train OD-VAE more efficiently, and a novel inference strategy is proposed to enable OD-VAE to handle videos of arbitrary length with limited GPU memory.
Comprehensive experiments on video reconstruction and LVDM-based video generation demonstrate the effectiveness and efficiency of our proposed methods.\footnote{Code: \href{https://github.com/PKU-YuanGroup/Open-Sora-Plan}{https://github.com/PKU-YuanGroup/Open-Sora-Plan}} \footnote{Equal Contribution: * ; Corresponding author: \dag}

\end{abstract}

\section{Introduction}
\label{sec:1}
Video generation has gained significant attention in both academia and industry, especially after the announcement of OpenAI's SORA \cite{videoworldsimulators2024}.
Currently,  Latent Video Diffusion Models (LVDMs), such as MagicTime \cite{magictime}, VideoComposer \cite{wang2024videocomposer}, AnimateDiff \cite{guoanimatediff}, Stable Video Diffusion (SVD) \cite{blattmann2023stable}, HiGen \cite{qing2024hierarchical}, Latte \cite{ma2024latte}, SORA \cite{videoworldsimulators2024}, Open-Sora \cite{opensora}, Open-Sora-Plan \cite{pku_yuan_lab_and_tuzhan_ai_etc_2024_10948109}, have been the dominators in video generation for their stability, effectiveness, and scalability.
These LVDMs share the same workflow: Variational Autoencoders (VAEs) \cite{kingma2014auto} compress origin videos into latent representations.
Then, the denoisers are trained to predict the noise added to these compressed representations.

However, the most frequently used VAE by LVDMs, Stable Diffusion VAE (SD-VAE) \cite{rombach2022high, podell2023sdxl}, is initially designed for spatially compressing images instead of videos.
When compressing a video, it treats each frame as an individual image, completely ignoring the redundancy in the temporal dimension.
This results in temporally redundant latent representation, which increases the input size of the following denoisers, leading to great hardware consumption for LVDMs.
In addition, the frame-wise compression of a video ignores the temporal information beneficial to reconstruction, causing lower reconstruction accuracy and reducing the quality of LVDMs' generated results.
Although the exploitation of temporal information is considered in the decoder of Stable Video Diffusion VAE (SVD-VAE) \cite{blattmann2023stable}, its compression of videos in the temporal dimension remains absent, which still brings a great hardware burden to LVDMs.

Furthermore, temporal compression for videos has been explored in some works about autoregressive-based video generation \cite{yan2021videogpt, ge2022long,yu2023magvit, yu2023language}.
They utilize VQ-VAEs \cite{van2017neural} to temp-spatially compress videos into discrete tokens and the following transformers are learned to predict these tokens.
Although these VQ-VAEs can't provide continuous latent representations for LVDMs, they still indicate the feasibility of temporal compression in LVDMs' VAEs.

To relieve the hardware burden of LVDMs and enhance their video generation ability with limited resources, we propose an omni-dimensional compression VAE (OD-VAE), which can temporally and spatially compress videos into concise latent representations.
Since a high temporal correlation exists in video frames, by strong 3D-Causal-CNN architecture \cite{yu2023language}, our OD-VAE can reconstruct video accurately with additional temporal compression. 
The sufficient compression and effective reconstruction of OD-VAE will greatly improve the efficiency of LVDMs.
To achieve a better trade-off between video reconstruction quality and compression speed, significant to the video generation results of LVDMs and their training speeds, respectively, we introduce and analyze four model variants of OD-VAE.
To train our OD-VAE more efficiently, we propose a novel tail initialization to exploit the weight of SD-VAE.
Besides, we propose novel temporal tiling, a split but one-frame overlap inference strategy, enabling OD-VAE to handle videos of arbitrary length with limited GPU memory.
\begin{figure*}[h]
\centering
\includegraphics[width=1.0\textwidth]{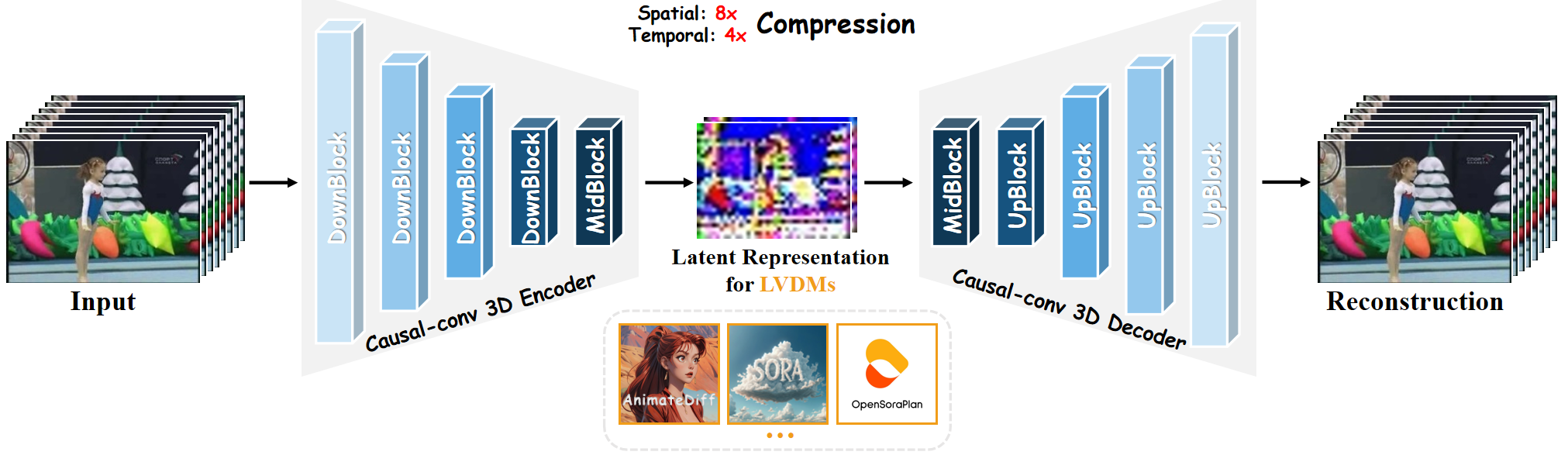}
\caption{The overview of our OD-VAE.
It adopts 3D-causal-CNN architecture to temp-spatially compress videos into concise latent representations and can reconstruct them accurately.
This greatly enhances the efficiency of LVDMs.
}
\label{fig1}
\end{figure*}

Our contributions are summarized as follows:
\begin{itemize}
    \item We propose OD-VAE, an omni-dimensional video compressor with a high reconstructed accuracy, which improves the efficiency of LVDMs.
    \item To achieve a better trade-off between video reconstruction quality and compression speed, we introduce and analyze four model variants of OD-VAE.
    \item To further improve the training efficiency and inference ability of our OD-VAE, we propose novel tile initialization and temporal tiling, respectively.
    \item Extensive experiments and ablations on video reconstruction and LVDM-based Video generation demonstrate the effectiveness and efficiency of our methods.
\end{itemize}

\section{Related Work}
\label{sec:2}

\subsection{Latent Video Diffusion Model}
\label{sec:2.1}
Latent Video Diffusion Models (LVDMs) is a significant task in artificial intelligence \cite{Prompt2Poster}.
It first use VAEs to compress videos into latent representations and then utilize denoisers to predict the noise added to them, have been developing rapidly since last year.
The OpenAI's SORA \cite{videoworldsimulators2024} that can generate videos of 1080P resolution and one minute long, greatly shocks the world.
LVDMs can be divided into two kinds in terms of the structures of their denoisers.
The first kind uses U-net-based denoisers \cite{ronneberger2015u, ChronoMagic-Bench, clh2023end}, such as MagicTime \cite{magictime}, AnimateDiff \cite{guoanimatediff}, and Stable Video Diffusion (SVD) \cite{blattmann2023stable}.
While the second kind utilizes Transformer-based denoisers \cite{peebles2023scalable}, such as Latte \cite{ma2024latte}, SORA \cite{videoworldsimulators2024}, Open-Sora \cite{opensora}, and Vidu \cite{bao2024vidu}.
Whatever the structures of the denoisers are, the VAEs determine the sizes of the inputs to denoisers and the reconstructed accuracy from latent representations to videos.
Thus, VAEs that provide concise representation while maintaining high reconstruction quality will greatly improve the efficiency of LVDMs.

\subsection{Variational Autoencoder}
\label{sec:2.2}
Variational Autoencoder (VAE) is initially designed for generation tasks by maximizing the Evidence Lower Bound (ELBO) of date\cite{kingma2014auto}.
Gradually, it has become a preceding component of other generation models and can be divided into two types.
The first is VQ-VAEs \cite{van2017neural}, which compress videos into discrete tokens and are used by autoregressive-based video generation models \cite{ge2022long, yu2023magvit, yu2023language}.
In these VQ-VAEs, temporal compressions for videos have existed, and the 3D-causal-CNN-based MAGVIT-v2 \cite{yu2023language} achieves state-of-the-art video reconstruction.
However, the discrete representations provided by VQ-VAEs are unsuitable for LVDMs.
The second is continuous VAEs, which compress videos into continuous representations and are used by LVDMs.
Among them, Stable Diffusion VAE (SD-VAE) \cite{rombach2022high}, and its decoder enhancement version, Stable Video Diffusion VAE (SVD-VAE) \cite{blattmann2023stable}, are the most popular. 
However, they only spatially compress videos while ignoring the temporal redundancy of videos.
Besides, we have discovered two works that are concurrent
with ours.
One is OPS-VAE \cite{opensora}, which utilizes two cascading VAEs to spatially and temporally compress videos, respectively.
The other is CV-VAE \cite{zhao2024cv}, which proposes a temporally compressed VAE but focuses more on latent space alignment to SD-VAE.
We will comprehensively compare our OD-VAE and them in the experiment.

\section{Method}
\label{sec:3}
In this section, we first provide the overview of OD-VAE, shown in Fig. \ref{fig1}.
Then, we discuss the four model variants of OD-VAE, shown in Fig. \ref{fig2}.
Finally, we introduce the tail initialization and temporal tiling.

\subsection{Overview of OD-VAE}
\label{sec:3.1}
Our OD-VAE adopts 3D-causal-CNN architecture to temporally and spatially compress videos into concise latent representations and can reconstruct them accurately, as shown in Fig. \ref{fig1}.
Since the structure of SD VAE is mature and stable, the basic design of our 3D-causal-CNN architecture is derived from it, which will be introduced in the next subsection.
Let $\mathcal{E}$ and $\mathcal{D}$ denote the encoder and the decoder of our OD-VAE, respectively.
A video containing $N+1$ frames is denoted as $\boldsymbol{X}=[\boldsymbol{x_1},\boldsymbol{x_2},...,\boldsymbol{x_{N+1}}] \in\mathcal{R}^{(N+1)\times H\times W\times 3}$, and the $i$-th frame of $\boldsymbol{X}$ is expressed as $\boldsymbol{x_i} \in \mathcal{R}^{H\times W\times 3}$.
The compressed latent representation of $\boldsymbol{X}$ is denoted as $\boldsymbol{Z} \in \mathcal{R}^{(n+1)\times h\times w\times c}$. 
When processing the video $\boldsymbol{X}$, OD-VAE keeps the temporal independence of its first frame $\boldsymbol{x_1}$, and only spatially compresses it.
In contrast, the following frames $\boldsymbol{x_i} (i>1)$ will be compressed in both the temporal and spatial dimensions.
This can be formulated as:
\begin{equation}
    \boldsymbol{Z} = \mathcal{E}(\boldsymbol{X})\text{.}
\end{equation}
The reconstruction is the inverse of the compression.
We use $\boldsymbol{\hat{X}} \in \mathcal{R}^{(N+1)\times H\times W\times 3}$ to express the reconstructed video and the process can be formulated as:
\begin{equation}
    \boldsymbol{\hat{X}} = \mathcal{D}(\boldsymbol{Z})\text{.}
\end{equation}
The temporal and spatial compression rates of OD-VAE are $c_t=\frac{n}{N}$ and $c_s=\frac{h}{H}=\frac{w}{W}$, respectively.
We set $c_s=8$ following SD-VAE and find $c_t=4$ will be a good trade-off between sufficient compression and accurate reconstruction.

\subsection{Model Variants of OD-VAE}
\label{sec:3.2}
Since video compression is necessary for the training of LVDMs, increasing the compression speed of our OD-VAE can greatly improve their training efficiency.
Hence, we introduce and analyze four different model variants of our OD-VAE, aiming to achieve a better trade-off between the compression speed and video reconstruction quality.

\begin{figure*}[h]
\centering
\includegraphics[width=1.0\textwidth]{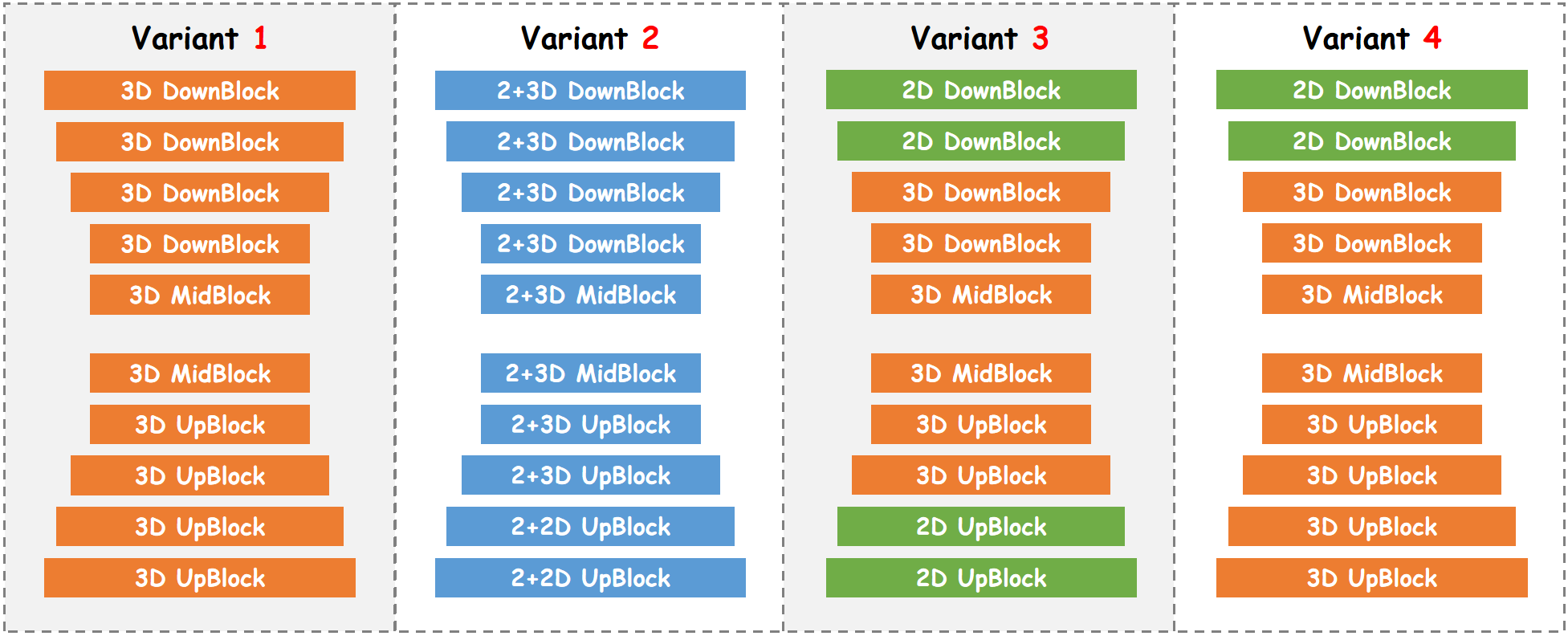}
\caption{Four variants of our OD-VAE.
\textbf{Variant 1:} inflating all the 2D convolutions in SD VAE to 3D convolutions.
\textbf{Variant 2:} replacing half of the 3D convolutions in variant 1 with 2D convolutions.
\textbf{Variant 3:} replacing the 3D convolutions in the outer blocks of variant 1's encoder and decoder with 2D convolutions.
\textbf{Variant 4:} replacing the 3D convolutions in the outer blocks of variant 1's encoder with 2D convolutions.}
\label{fig2}
\end{figure*}

\textbf{Variant 1.}
An easy way to extend SD VAE to our 3D-causal-CNN-based OD-VAE is inflating all the 2D convolutions into 3D convolutions by adding a temporal dimension to all 2D kernels, shown in Fig. \ref{fig2}(a).
The video reconstruction ability of variant 1 is the best since its full-3D architecture can completely exploit the temporal and spatial information in the video by making features temp-spatially interact at each convolution.
However, numerous expensive 3D convolutions in the network lead to a slow compression speed, lowering the training efficiency of LVDMs.

\textbf{Variant 2.}
Since numerous 3D convolutions in variant 1 lead to a slow compression speed, we utilize an intuitive way to reduce expensive 3D convolutions.
Specifically, we replace half of the 3D convolutions in variant 1 with 2D convolutions and obtain variant 2, shown in Fig. \ref{fig2}(b).
In variant 2, half of its convolutions are limited to only conducting spatial transformation for the input features, lowering the computational consumption of compression.
As half of the convolutions can still process the features omni-dimensionally, abundant temporal and spatial information in a video is still well utilized, guaranteeing its reconstruction ability.

\textbf{Variant 3.}
However, in variant 1, the consumption of each 3D convolution is different.
The 3D convolutions in the outer blocks process large-sized features with huge expense while those in the inner blocks process small-sized features with little expense.
Hence, replacing a 3D convolution in an outer block leads to a greater reduction in consumption than replacing one in an inner block.
Based on this, we utilize a more reasonable replacement strategy for variant 1 and obtain variant 3.
Specifically, we replace all the 3D convolutions in some outer blocks with 2D convolutions while maintaining the other inner blocks unchanged, shown in Fig. \ref{fig2}(c).
With this strategy, the compression speed of variant 3 will probably be faster than that of variant 2.

\textbf{Variant 4.}
Since the decoder of OD-VAE doesn't participate in video compression, the convolution replacement in the decoder can't improve the training efficiency of LVDMs while lowering the reconstruction accuracy.
Therefore, we keep the decoder of variant 1 unchanged and only replace the 3D convolutions in the outer blocks of the encoder with 2D convolutions and obtain variant 4, shown in Fig. \ref{fig2}(d).
With a full 3D decoder, the video reconstruction ability of variant 4 will probably be better than that of variant 3.

\subsection{Tail Initialization and Temporal Tiling}
\label{sec:3.3}
\textbf{Tail Initialization.}
Notably, when $N=0$, the video $\boldsymbol{X}$ degrades as an image and our OD-VAE can be viewed as an image VAE.
This brings the potential for OD-VAE to inherit the spatial compression and reconstruction ability of powerful SD VAE.
With this inheritance of ability in the spatial dimension, the training efficiency of our OD-VAE is higher, since the spatial prior will accelerate the convergence of our model.
Hence, for better inheritance, we design a special initialization method to utilize the weight of 2D SD-VAE perfectly, named tail initialization.
Specifically, we denote a 5 dimension 3D convolution kernel in the OD-VAE as $\boldsymbol{K_{3D}} \in \mathcal{R}^{I\times O\times T\times H\times W}$, and its corresponding 4 dimension 2D kernel in SD VAE as $\boldsymbol{K_{2D}} \in \mathcal{R}^{I\times O\times H\times W}$.
For $\boldsymbol{K_{3D}}$, we use the weight of $\boldsymbol{K_{2D}}$ to initial its temporally last element and set other elements to $0$, expressed as:
\begin{equation}
    \boldsymbol{K_{3D}}[:,:,i,:,:] = \begin{cases}  
\boldsymbol{K_{2D}}, & \text{if } i = -1\text{.} \\  
\boldsymbol{0}, & \text{else.}  
\end{cases}
\end{equation}
We use $\boldsymbol{F_{3D}}$ and $\boldsymbol{F_{2D}}$ to denote the input feature maps of $\boldsymbol{K_{3D}}$ and $\boldsymbol{K_{2D}}$, respectively.
With tail initialization, before training, our OD-VAE satisfies the following equation:
\begin{equation}
    \boldsymbol{F_{3D}}*\boldsymbol{K_{3D}}= \boldsymbol{F_{2D}}*\boldsymbol{K_{2D}}\text{.}
\end{equation}
The equation means that our OD-VAE can compress an image into a latent representation and reconstruct it accurately as SD-VAE without learning.
This indicates that the spatial compression and reconstruction ability of SD-VAE is completely transferred to our OD-VAE.
The strong spatial prior accelerates the convergence of our OD-VAE, greatly enhancing the training efficiency.

\textbf{Temporal Tiling.}
Since long video generation has been a main trend, enabling our OD-VAE to handle videos of arbitrary length with limited GPU memory is necessary.
Hence, we design a split but one-frame overlap inference strategy, named temporal tiling.
Specifically, we temporally split a video $X$ into $M$ groups, denoting as $[\boldsymbol{X_1},\boldsymbol{X_2},...,\boldsymbol{X_M}]$.
The last frame of $\boldsymbol{X_i}$ and the first frame of $\boldsymbol{X_{i+1}}$ are the same.
We compress each group $\boldsymbol{X_i}$ into latent representation $\boldsymbol{Z_i}$ individually.
Then, we drop the first frames of $\boldsymbol{Z_i}$ when $i>1$ and concatenate $\boldsymbol{Z_i}(1\leq i\leq M)$ along temporal dimension to obtain $\boldsymbol{Z}$.
We introduce the same grouping mechanism to the reconstructed video $\boldsymbol{\hat{X}}$ that $\boldsymbol{\hat{X}}=[\boldsymbol{\hat{X}_1},\boldsymbol{\hat{X}_2},...,\boldsymbol{\hat{X}_M}]$.
To reconstruct $\boldsymbol{Z}$ as $\boldsymbol{\hat{X}}$, we first decode $\boldsymbol{Z_i}$ into $\boldsymbol{\hat{X}_i}$ individually.
Then, we drop the first frames of $\boldsymbol{\hat{X}_i}$ when $i>1$ and concatenate $\boldsymbol{\hat{X}_i}(1\leq i\leq M)$ along temporal dimension.
As a high temporal correlation exists in video frames, the overlap can connect each group well and greatly reduce compressed and reconstructed errors.

\section{Experiment}
\label{sec:4}
In this section, we first introduce the experimental setting, including models, training strategy, and evaluation details.
Then, comprehensive comparisons between OD-VAE and other baselines on video reconstruction and LVDM-based video generation are conducted to demonstrate the superiority of our OD-VAE.
Finally, extensive ablations are provided to certify the effectiveness of our proposed methods.

\begin{figure*}[h]
\centering
\includegraphics[width=1.0\textwidth]{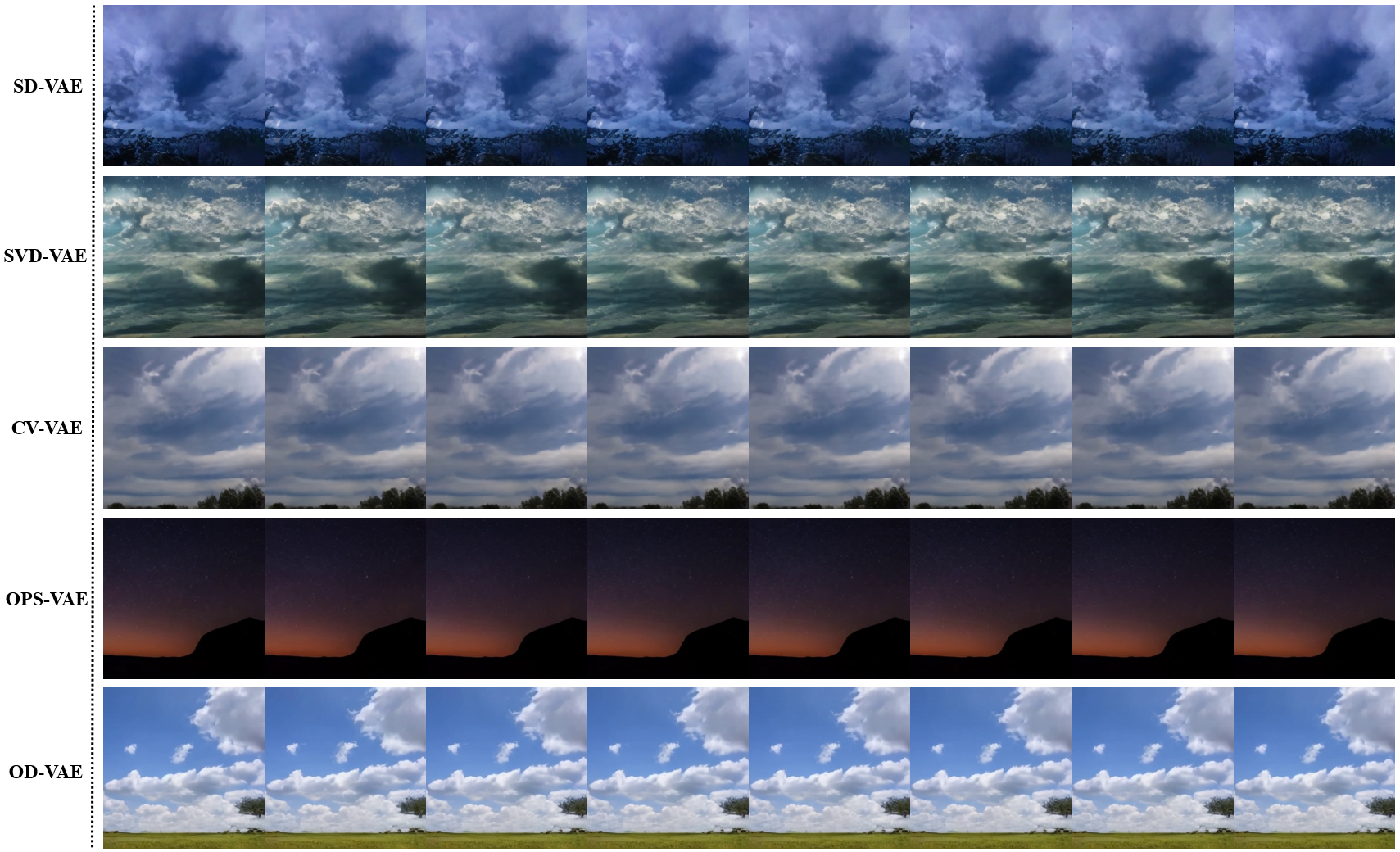}
\caption{Video generation results of LVDMs with different VAEs on the SkyTimelapse dataset.
As the figure shows, with OD-VAE, LVDM can generate more realistic and high-quality videos.}
\label{fig3}
\end{figure*}

\subsection{Experimental Setting}
\label{sec:4.1}
\textbf{Models.}
To demonstrate the effectiveness and efficiency of our OD-VAE, we compare it with six other state-of-the-art commonly used VAEs in terms of video reconstruction and LVDM-based video generation, including:
(1) VQGAN \cite{esser2021taming}: a widely used image VQ-VAE.
Following \cite{zhao2024cv}, we use its f8-8192 version in our experiment.
(2) TATS \cite{ge2022long}: a 3D video VQ-VAE applied to autoregressive-based video generation.
(3) SD-VAE \cite{rombach2022high}: the most frequently used image VAE by LVDMs.
Following \cite{zhao2024cv}, we use its numerically stable version, SD2.1-VAE.
(4) SVD-VAE \cite{blattmann2023stable}: A video VAE obtained by enhancing the decoder of SD-VAE.
It shares the same encoder structure as SD-VAE.
(5) CV-VAE \cite{zhao2024cv}: a video VAEs contemporaneous with our research.
(6) OPS-VAE \cite{opensora}: another video VAEs also contemporaneous with our research.
It first conducts spatial downsample then temporal downsample to an input video.
As discrete VQGAN and TATS aren't suitable for LVDMs, they are only used for experiments on video reconstruction. 
In the method section, we introduce four model variants of our OD-VAEs.
We use variant 4 of our OD-VAE to compare to other baselines, since according to the ablations, variant 4 achieves the best trade-off between the video reconstruction quality and compression speed among all the variants.

\textbf{Training strategy.}
We use Adma optimizer \cite{kingma2014adam} to train our OD-VAE for 650k steps, with a constant learning rate $1\times 10^{-5}$ and batch size 8.
The training dataset of our OD-VAE contains 440k self-scrape internet videos and 220k videos from the K400 dataset \cite{carreira2017quo}.
During training, all the input videos are processed to clips of 25-frame length and $256\times 256$ resolution.
Following \cite{esser2021taming,podell2023sdxl}, the loss function contains a reconstruction term, a KL term, and an adversarial term \cite{goodfellow2014generative}.
To obtain more stable training results, following \cite{peebles2023scalable, podell2023sdxl}, we utilize an exponential moving average (EMA) of OD-VAE weights over training with a decay of 0.999.
Since SD2.1-VAE is numerically stable, We use its weights to initialize our OD-VAE, enhancing the training efficiency.
The training is conducted on 8 NVIDIA 80G A100 GPUs with Pytorch \cite{paszke2019pytorch}. 

 \begin{table*}[htbp]
 \centering
 \caption{Video reconstruction results of VAEs on the WebVid-10M validation set and Panda-70M validation set, along with their numbers of parameters and video compression rate.
 For video reconstruction metrics, with the highest video compression rate, the best and second scores are indicated in \textbf{bold} and \underline{underlined}.}
 \scalebox{1.00}{
\begin{tabular}{c|cc|ccc|ccc}
 \hline
   \multirow{2}{*}{Method}  & \multirow{2}{*}{VCPR} & \multirow{2}{*}{Params} & \multicolumn{3}{c}{WebVid-10M}& \multicolumn{3}{|c}{Panda-70M}\\
   \cline{4-9}
   &&&PSNR$(\uparrow)$ &SSIM$(\uparrow)$ &LPIPS$(\downarrow)$ &PSNR$(\uparrow)$ &SSIM$(\uparrow)$ &LPIPS$(\downarrow)$\\
\hline
VQGAN&\multirow{3}{*}{$64(1\times8\times8)$}&69.00M&26.26&0.7699&0.0906&26.07&0.8295&0.0722\\
SD-VAE&&83.65M&30.19&0.8379&0.0568&30.40&0.8894&0.0396\\
SVD-VAE&&97.74M&31.15&0.8686&0.0547&31.00&0.9058&0.0379\\
\hline
TATS&\multirow{4}{*}{$256(4\times8\times8)$}&52.19M&23.10&0.6758&0.2645&21.77&0.6680&0.2858\\
CV-VAE&&182.45M&30.76&0.8566&\underline{0.0803}&29.57&0.8795&0.0673\\
OPS-VAE&&393.34M&\underline{31.12}&\underline{0.8569}&0.1003&\textbf{31.06}&\underline{0.8969}&\underline{0.0666}\\
OD-VAE&&239.19M&\textbf{31.16}&\textbf{0.8694}&\textbf{0.0586}&\underline{30.49}&\textbf{0.8970}&\textbf{0.0454}\\
\hline
 \end{tabular}}
\label{Table1} 
 \end {table*}

\begin{table*}[htbp]
 \centering
 \caption{Video generation results and the training efficiency of LVDMs with different VAEs on the UCF101 and SkyTimelapse dataset.
 The best and second scores are indicated in \textbf{bold} and \underline{underlined} for all these metrics.}
 \scalebox{1.00}{
 \begin{tabular}{c|ccccc|cccc}
 \hline
   \multirow{2}{*}{Method}  & \multicolumn{5}{c|}{UCF101}& \multicolumn{4}{c}{SkyTimelapse}\\
   \cline{2-10}
   &FVD$(\downarrow)$ &KVD$(\downarrow)$ &IS$(\uparrow)$ &TMem$(\downarrow)$ &TSpeed$(\uparrow)$ &FVD$(\downarrow)$ &KVD$(\downarrow)$ &TMem$(\downarrow)$ &TSpeed$(\uparrow)$\\
\hline
SD-VAE&1685.29&116.10&33.00&74364MB&0.87it/s&325.00&26.28&74498MB&0.86it/s\\
SVD-VAE&1663.98&\textbf{108.27}&31.41&74364MB&0.87it/s&\textbf{285.23}&25.10&74498MB&0.86it/s\\
CV-VAE&\underline{1380.43}&129.29&\textbf{61.11}&\underline{30628MB}&1.31it/s&326.86&\underline{23.57}&\underline{30938MB}&\underline{1.29it/s}\\
OPS-VAE&1502.64&142.4&53.13&31220MB&\underline{1.52it/s}&312.22&24.47&31516MB&1.49it/s\\
\hline
OD-VAE&\textbf{1315.13}&\underline{110.88}&\underline{58.98}&\textbf{30520MB}&\textbf{1.80it/s}&\underline{294.31}&\textbf{20.76}&\textbf{30834MB}&\textbf{1.76it/s}\\
\hline
 \end{tabular}}
\label{Table2} 
 \end {table*}

\textbf{Evaluation details.}
For evaluation on video reconstruction, we select two popular large open-domain video datasets, WebVid-10M \cite{bain2021frozen} and Panda-70M \cite{chen2024panda}.
we only use their validation sets for efficiency and fairness.
For each video in these two validation sets, we transform it to a clip of 25-frame length and $256\times 256$ resolution.
To quantify models' video reconstruction ability, we use three popular metrics, peak signal-to-noise ratio (PSNR) \cite{hore2010image}, structural similarity index measure (SSIM) \cite{wang2004image}, and Learned Perceptual Image Patch Similarity (LPIPS) \cite{zhang2018unreasonable}.
We also use the video compression rate (VCPR) and the number of parameters (Params) to denote the video compression level and the network complexity of these VAEs, respectively.
To evaluate these VAEs' effect on LVDM-based video generation, we fix the structure of the denoiser and change its previous VAE.
We select Latte's denoiser \cite{ma2024latte}, since it uses a novel SORA-like transformer-based structure and achieves excellent results in LVDM-based video generation.
Following \cite{ge2022long}, we choose two public datasets, UCF101 \cite{soomro2012dataset} and SkyTimelapse \cite{xiong2018learning}, for class-conditional and unconditional generation respectively.
We use almost the same setting introduced in \cite{ma2024latte} to train Latte's denoiser with these VAEs for 200k steps.
The only difference is that we use longer video clips of 81-frame length and adjust the batch size to fit the memory limitation of a single GPU.
To assess the quality of the generated videos on the two datasets, we employ two popular metrics, Frechet Video Distance (FVD) and Kernel Video Distance (KVD) \cite{unterthiner2018towards}.
In addition, we also report models' Inception Score (IS) \cite{saito2020train} on the UCF101 dataset, calculated by a trained C3D model \cite{tran2015learning}.
These metrics are calculated based on 2048 samples.
To measure LVDM's efficiency with different VAEs, we list their training GPU memory consumption (TMem) and training speed (TSpeed) on the two datasets.
These tests are conducted on NVIDIA 80G A100 GPUs.

\begin{table}[t]
 \centering
 \caption{The compression speed (CSpeed) of the four variants and the training speed (TSpeed) of LVDM with them.}
 \scalebox{0.95}{
 \begin{tabular}{ccccc}
 \hline
   Model &Variant 1 &Variant 2 &Variant 3&Variant 4 \\
\hline
CSpeed($\uparrow$) &2.15it/s&2.44it/s&4.13it/s&4.13it/s\\
TSpeed($\uparrow$)
&1.26it/s&1.38it/s&1.80it/s&1.80it/s \\
\hline
 \end{tabular}}
\label{Table3} 
 \end {table}
\begin{figure*}[htbp]
\centering
\subfigure[PSNR of the four variants.]{
\begin{minipage}[t]{0.323\linewidth}
\centering
\includegraphics[width=2.175in]{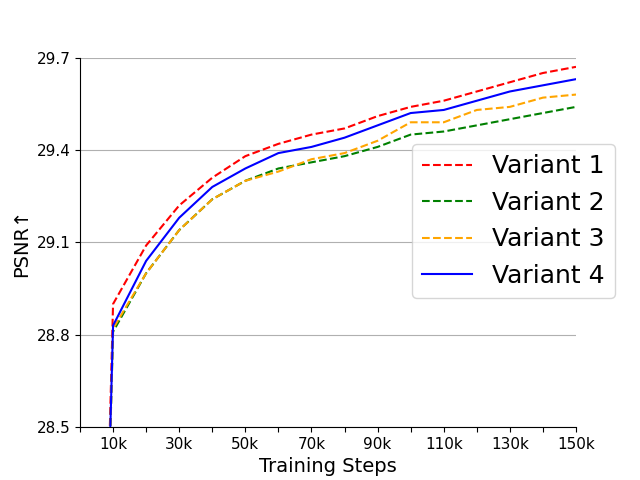}
\end{minipage}
}%
\subfigure[LPIPS of the four variants.]{
\begin{minipage}[t]{0.323\linewidth}
\centering
\includegraphics[width=2.175in]{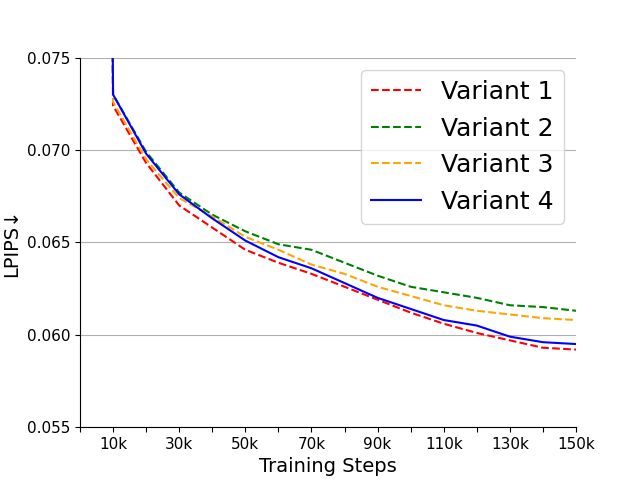}
\end{minipage}
}%
\subfigure[FVD of the four variants.]{
\begin{minipage}[t]{0.323\linewidth}
\centering
\includegraphics[width=2.175in]{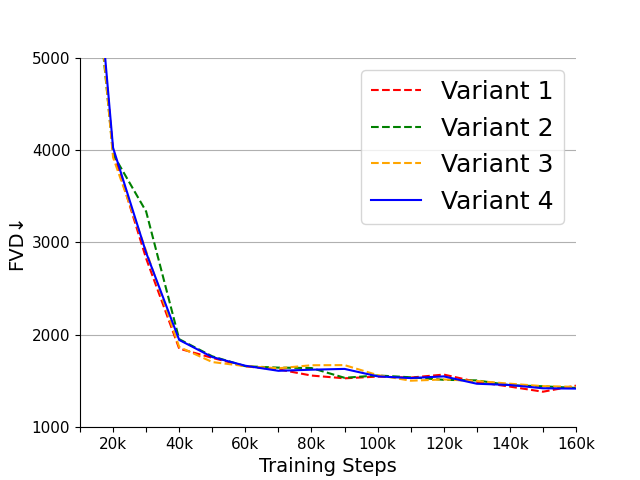}
\end{minipage}
}

\subfigure[PSNR of the three initialization methods.]{
\begin{minipage}[t]{0.323\linewidth}
\centering
\includegraphics[width=2.175in]{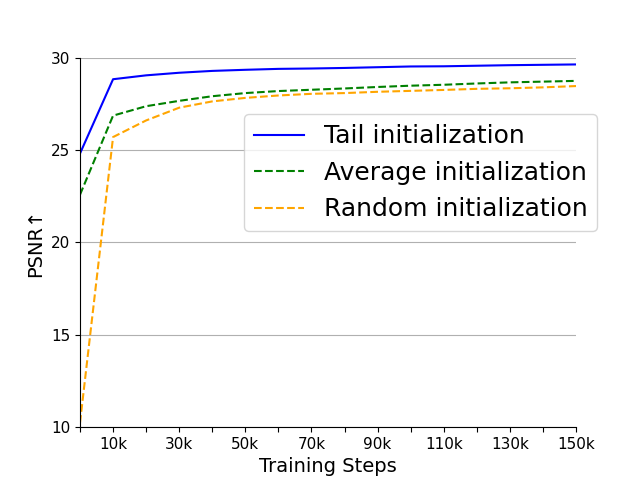}
\end{minipage}
}%
\subfigure[LPIPS of the three initialization methods.]{
\begin{minipage}[t]{0.323\linewidth}
\centering
\includegraphics[width=2.175in]{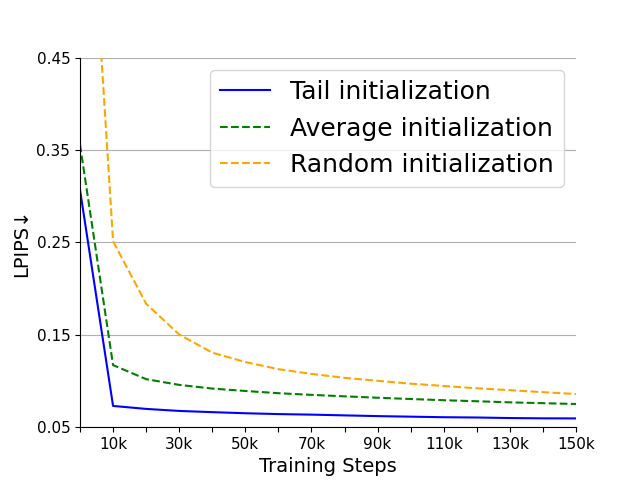}
\end{minipage}
}%
\subfigure[FVD of the three initialization methods.]{
\begin{minipage}[t]{0.323\linewidth}
\centering
\includegraphics[width=2.175in]{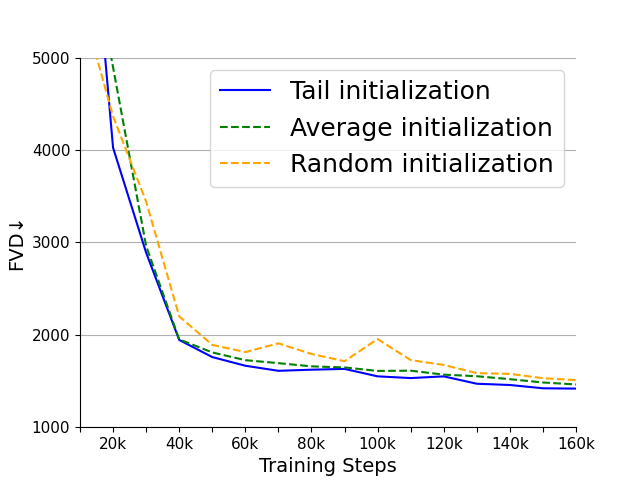}
\end{minipage}
}
\centering
\caption{(a), (b) are the PSNR and LPIPS of the four variants on the WebVid-10M validation set.
(c) is the FVD of the four variants on the UCF101 dataset.
(d), (e) are the PSNR and LPIPS of the three initialization methods on the WebVid-10M validation set.
(f) is the FVD of the three initialization methods on the UCF101 dataset.
}
\label{fig4}
\end{figure*}

\subsection{Comparison with Other Baselines}
\label{sec:4.2}
We display the video reconstruction results of our OD-VAE and other baselines in Table. \ref{Table1}.
The results in Table. \ref{Table1} reflect that although OD-VAE can $4\times$ temporally compress videos, its reconstruction quality is not inferior to commonly used SD-VAE and SVD-VAE.
For example, the PSNR and SSIM of our OD-VAE on the WebVid-10M validation set are 0.97 and 0.0315 higher than that of SD-VAE, respectively.
Compared to SVD-VAE, although the overall performance of our OD-VAE is worse, its PSNR and SSIM on the WebVid-10M validation set are still slightly higher.
This proves that our OD-VAE can fully exploit the temporal redundancy of video frames to obtain a more concise latent representation while maintaining high reconstructed quality.
Furthermore, our OD-VAE behaves better than the two works concurrent with us, CV-VAE and OPS-VAE, which proves the effectiveness of our model design and training strategy.
For example, the SSIM of OD-VAE on the WebVid-10M validation set is 0.0128 and 0.0125 higher than that of CV-VAE and OPS-VAE, respectively.
On the Panda-70M validation set, the LPIPS of our OD-VAE is 0.0219 and 0.0211 lower than that of CV-VAE and OPS-VAE, respectively.

In Table. \ref{Table2}, we display the LVDM-based video generation results of our OD-VAE and other baselines.
The results in Table. \ref{Table2} show that, through $4\times$ temporal compression of VAEs, the efficiency of LVDM is greatly improved.
On the two datasets, the video generation results of our OD-VAE are better than that of SD-VAE and SVD-VAE, while the training consumption is greatly reduced. 
For example, on the UCF101 dataset, with the same training steps, using our OD-VAE can achieve better FVD (370.16 lower than that of SD-VAE and 348.85 lower than that of SVD-VAE) and faster training speed ($2.06\times$ that of SD-VAE and SVD-VAE).
Furthermore, compared to CV-VAE and OPS-VAE, although the video compression rate is the same, our OD-VAE brings better video generation results and lower training consumption to LVDM.
For example, on the SkyTimelapse dataset, with the same training steps, using our OD-VAE can obtain better FVD (32.55 lower than that of CV-VAE and 17.91 lower than that of OPS-VAE) and faster training speed (0.47it/s faster than that of SD-VAE and 0.27it/s faster than that of OPS-VAE).
Besides, we show some visual results of LVDM with different VAEs on the SkyTimelapse dataset in Fig. \ref{fig3}.
According to Fig. \ref{fig3}, with OD-VAE, LVDM can generate more realistic and high-quality videos.

\subsection{Ablation Experiment}
\label{sec:4.3}
\textbf{Model variant.}
To obtain the variant with the best trade-off between video reconstruction quality and compression speed, we train the four model variants of OD-VAE for 150k steps with the same setting mentioned above.
We show their PSNR and LPIPS on the WebVid-10M validation set in Fig. \ref{fig4} (a) and (b).
Besides, we use their final checkpoints to train Latte's denoiser on the UCF101 dataset with the same setting mentioned above and report their FVD in Fig. \ref{fig4} (c).
The compression speed (CSpeed) of the four variants, calculated by processing videos of 81-frame length and $256\times 256$ resolution, along with the training speed (TSpeed) of LVDM with them on the UCF101 dataset, are listed in Table. \ref{Table3}.
According to Fig. \ref{fig4} (a), (b), the PSNR and SSIM of variant 4 are slightly worse than that of variant 1 but better than the other variants.
Since the reconstruction abilities of the four variants are close, using them as the preceding components of LVDM causes similar results of video generation, shown in Fig. \ref{fig4} (c).
However, according to the Table. \ref{Table3}, the compression speed of variant 4 is much faster than that of variant 1 and variant 2, bringing extreme efficiency enhancement to the training of LVDM.
Hence, our OD-VAE utilizes variant 4 as the final structure, achieving the best trade-off between video reconstruction quality and compression speed.

\textbf{Initialization Method.}
To verify the effectiveness of our tail initialization, we compare it with two other initialization methods, average initialization, and random initialization.
Average initialization can be expressed as:
\begin{equation}
    \boldsymbol{K_{3D}}[:,:,i,:,:] = \frac{\boldsymbol{K_{2D}}}{T}( 1\leq i\leq T)\text{.}
\end{equation}
The random initialization means we randomly initialize our OD-VAE with Gaussian random numbers.
We initialize our OD-VAE with the three methods and train the three versions for 150k steps with the same setting mentioned above, respectively.
We show their PSNR and LPIPS on the WebVid-10M validation set in Fig. \ref{fig4} (d) and (e).
Besides, we use their final checkpoints to train Latte’s denoiser on the UCF101 dataset with the same setting mentioned above and report their FVD in Fig. \ref{fig4} (f). 
According to Fig. \ref{fig4} (d), (e), and (f), with the same training steps, using tail initialization can greatly improve the video reconstruction ability of our OD-VAE and the video generation quality of LVDM.

\begin{table}[t]
 \centering
 \caption{The PSNR, LPIPS of our OD-VAE (w or w/o temporal tiling) on the WebVid-10 validation set and the FVD, IS of corresponding LVDM on the UCF101 dataset.}
 \scalebox{1.00}{
 \begin{tabular}{c|cc|cc}
 \hline
   Temporal& \multicolumn{2}{|c}{WebVid-10M}& \multicolumn{2}{|c}{UCF101}\\
\cline{2-5}
   Tiling&PSNR$\uparrow$ &LPIPS$\downarrow$ &FVD$\downarrow$ &IS$\uparrow$ \\
\hline
$\boldsymbol{\times}$ &31.05&0.0589&1315.13&58.98 \\
\checkmark &30.98&0.0591&1331.46&58.89\\
\hline
 \end{tabular}}
\label{Table4} 
 \end {table}

\textbf{Temporal Tiling}
When directly compressing and reconstructing a video of $256\times 256$ resolution on an NVIDIA 80G A100 GPU, the maximum length of frames our OD-VAE can process is 125.
With temporal tiling, our OD-VAE can handle a video in groups and the original length limitation disappears.
This enables LVDM to generate longer videos.
To evaluate the effect of temporal tiling on video reconstruction and LVDM-based video generation, we conduct experiments on the WebVid-10M validation set and the UCF101 dataset with the same setting mentioned above, respectively.
We fix the length of a group to 33 and increase the frame length of the WebVid-10M validation clips from 33 to 97.
In Table. \ref{Table4}, we list the PSNR and LPIPS on the WebVid-10M validation set, and the FVD and IS on the UCF101 dataset.
According to Table. \ref{Table4}, with temporal tiling, these metrics slightly decrease, which means temporal tiling will not do much harm to the video reconstruction ability of our OD-VAE and the video generation quality of corresponding LVDM. 

\section{Conclusion}
\label{sec:5}
In this work, we proposed a novel omni-dimensional compression VAE for improving LVDMs, termed OD-VAE.
It utilized effective 3D-causal-CNN architecture to $4\times$ temporally and $8\times$ spatially compress videos into latent representations while maintaining high reconstructed accuracy.
These more concise representations reduced the input size of LVDMs' denoisers, greatly improving the efficiency of LVDMs. 
To achieve a better trade-off between video reconstruction quality and compression speed, we introduced and analyzed four variants of our OD-VAE.
To train OD-VAE more efficiently, we proposed a novel tail initialization to exploit the weight of SD-VAE perfectly.
Besides, we proposed temporal tiling, a split but one-frame overlap inference strategy, enabling our OD-VAE to process videos of arbitrary length with limited GPU memory.
Comprehensive experiments and ablations on video reconstruction and LVDM-based video generation demonstrated the effectiveness and efficiency of our proposed methods.    

%

\small
\bibliographystyle{ieeenat_fullname}
\bibliography{main}

\begin{thebibliography}{39}
\providecommand{\natexlab}[1]{#1}
\providecommand{\url}[1]{\texttt{#1}}
\expandafter\ifx\csname urlstyle\endcsname\relax
  \providecommand{\doi}[1]{doi: #1}\else
  \providecommand{\doi}{doi: \begingroup \urlstyle{rm}\Url}\fi

\bibitem[Bain et~al.(2021)Bain, Nagrani, Varol, and Zisserman]{bain2021frozen}
Max Bain, Arsha Nagrani, G{\"u}l Varol, and Andrew Zisserman.
\newblock Frozen in time: A joint video and image encoder for end-to-end retrieval.
\newblock In \emph{Proceedings of the IEEE/CVF international conference on computer vision}, pages 1728--1738, 2021.

\bibitem[Bao et~al.(2024)Bao, Xiang, Yue, He, Zhu, Zheng, Zhao, Liu, Wang, and Zhu]{bao2024vidu}
Fan Bao, Chendong Xiang, Gang Yue, Guande He, Hongzhou Zhu, Kaiwen Zheng, Min Zhao, Shilong Liu, Yaole Wang, and Jun Zhu.
\newblock Vidu: a highly consistent, dynamic and skilled text-to-video generator with diffusion models.
\newblock \emph{arXiv preprint arXiv:2405.04233}, 2024.

\bibitem[Blattmann et~al.(2023)Blattmann, Dockhorn, Kulal, Mendelevitch, Kilian, Lorenz, Levi, English, Voleti, Letts, et~al.]{blattmann2023stable}
Andreas Blattmann, Tim Dockhorn, Sumith Kulal, Daniel Mendelevitch, Maciej Kilian, Dominik Lorenz, Yam Levi, Zion English, Vikram Voleti, Adam Letts, et~al.
\newblock Stable video diffusion: Scaling latent video diffusion models to large datasets.
\newblock \emph{arXiv preprint arXiv:2311.15127}, 2023.

\bibitem[Brooks et~al.(2024)Brooks, Peebles, Holmes, DePue, Guo, Jing, Schnurr, Taylor, Luhman, Luhman, Ng, Wang, and Ramesh]{videoworldsimulators2024}
Tim Brooks, Bill Peebles, Connor Holmes, Will DePue, Yufei Guo, Li Jing, David Schnurr, Joe Taylor, Troy Luhman, Eric Luhman, Clarence Ng, Ricky Wang, and Aditya Ramesh.
\newblock Video generation models as world simulators.
\newblock 2024.

\bibitem[Carreira and Zisserman(2017)]{carreira2017quo}
Joao Carreira and Andrew Zisserman.
\newblock Quo vadis, action recognition? a new model and the kinetics dataset.
\newblock In \emph{proceedings of the IEEE Conference on Computer Vision and Pattern Recognition}, pages 6299--6308, 2017.

\bibitem[Chen et~al.(2023)Chen, Wang, and Chen]{clh2023end}
Liuhan Chen, Yirou Wang, and Yongyong Chen.
\newblock End-to-end xy separation for single image blind deblurring.
\newblock In \emph{Proceedings of the 31st ACM International Conference on Multimedia}, pages 1273--1282, 2023.

\bibitem[Chen et~al.(2024)Chen, Siarohin, Menapace, Deyneka, Chao, Jeon, Fang, Lee, Ren, Yang, et~al.]{chen2024panda}
Tsai-Shien Chen, Aliaksandr Siarohin, Willi Menapace, Ekaterina Deyneka, Hsiang-wei Chao, Byung~Eun Jeon, Yuwei Fang, Hsin-Ying Lee, Jian Ren, Ming-Hsuan Yang, et~al.
\newblock Panda-70m: Captioning 70m videos with multiple cross-modality teachers.
\newblock In \emph{Proceedings of the IEEE/CVF Conference on Computer Vision and Pattern Recognition}, pages 13320--13331, 2024.

\bibitem[Esser et~al.(2021)Esser, Rombach, and Ommer]{esser2021taming}
Patrick Esser, Robin Rombach, and Bjorn Ommer.
\newblock Taming transformers for high-resolution image synthesis.
\newblock In \emph{Proceedings of the IEEE/CVF conference on computer vision and pattern recognition}, pages 12873--12883, 2021.

\bibitem[Ge et~al.(2022)Ge, Hayes, Yang, Yin, Pang, Jacobs, Huang, and Parikh]{ge2022long}
Songwei Ge, Thomas Hayes, Harry Yang, Xi Yin, Guan Pang, David Jacobs, Jia-Bin Huang, and Devi Parikh.
\newblock Long video generation with time-agnostic vqgan and time-sensitive transformer.
\newblock In \emph{European Conference on Computer Vision}, pages 102--118. Springer, 2022.

\bibitem[Goodfellow et~al.(2014)Goodfellow, Pouget-Abadie, Mirza, Xu, Warde-Farley, Ozair, Courville, and Bengio]{goodfellow2014generative}
Ian Goodfellow, Jean Pouget-Abadie, Mehdi Mirza, Bing Xu, David Warde-Farley, Sherjil Ozair, Aaron Courville, and Yoshua Bengio.
\newblock Generative adversarial nets.
\newblock \emph{Advances in neural information processing systems}, 27, 2014.

\bibitem[Guo et~al.()Guo, Yang, Rao, Liang, Wang, Qiao, Agrawala, Lin, and Dai]{guoanimatediff}
Yuwei Guo, Ceyuan Yang, Anyi Rao, Zhengyang Liang, Yaohui Wang, Yu Qiao, Maneesh Agrawala, Dahua Lin, and Bo Dai.
\newblock Animatediff: Animate your personalized text-to-image diffusion models without specific tuning.
\newblock In \emph{The Twelfth International Conference on Learning Representations}.

\bibitem[Hore and Ziou(2010)]{hore2010image}
Alain Hore and Djemel Ziou.
\newblock Image quality metrics: Psnr vs. ssim.
\newblock In \emph{2010 20th international conference on pattern recognition}, pages 2366--2369. IEEE, 2010.

\bibitem[Kingma and Ba(2015)]{kingma2014adam}
Diederik~P Kingma and Jimmy Ba.
\newblock Adam: A method for stochastic optimization.
\newblock \emph{International Conference on Learning Representations}, 2015.

\bibitem[Kingma and Welling(2014)]{kingma2014auto}
Diederik~P Kingma and Max Welling.
\newblock Auto-encoding variational bayes.
\newblock \emph{stat}, 1050:\penalty0 1, 2014.

\bibitem[Lab and etc.(2024)]{pku_yuan_lab_and_tuzhan_ai_etc_2024_10948109}
PKU-Yuan Lab and Tuzhan~AI etc.
\newblock Open-sora-plan, 2024.

\bibitem[Ma et~al.(2024)Ma, Wang, Jia, Chen, Liu, Li, Chen, and Qiao]{ma2024latte}
Xin Ma, Yaohui Wang, Gengyun Jia, Xinyuan Chen, Ziwei Liu, Yuan-Fang Li, Cunjian Chen, and Yu Qiao.
\newblock Latte: Latent diffusion transformer for video generation.
\newblock \emph{arXiv preprint arXiv:2401.03048}, 2024.

\bibitem[Paszke et~al.(2019)Paszke, Gross, Massa, Lerer, Bradbury, Chanan, Killeen, Lin, Gimelshein, Antiga, et~al.]{paszke2019pytorch}
Adam Paszke, Sam Gross, Francisco Massa, Adam Lerer, James Bradbury, Gregory Chanan, Trevor Killeen, Zeming Lin, Natalia Gimelshein, Luca Antiga, et~al.
\newblock Pytorch: An imperative style, high-performance deep learning library.
\newblock \emph{Advances in neural information processing systems}, 32, 2019.

\bibitem[Peebles and Xie(2023)]{peebles2023scalable}
William Peebles and Saining Xie.
\newblock Scalable diffusion models with transformers.
\newblock In \emph{Proceedings of the IEEE/CVF International Conference on Computer Vision}, pages 4195--4205, 2023.

\bibitem[Podell et~al.(2023)Podell, English, Lacey, Blattmann, Dockhorn, M{\"u}ller, Penna, and Rombach]{podell2023sdxl}
Dustin Podell, Zion English, Kyle Lacey, Andreas Blattmann, Tim Dockhorn, Jonas M{\"u}ller, Joe Penna, and Robin Rombach.
\newblock Sdxl: Improving latent diffusion models for high-resolution image synthesis.
\newblock \emph{arXiv preprint arXiv:2307.01952}, 2023.

\bibitem[Qing et~al.(2024)Qing, Zhang, Wang, Wang, Wei, Zhang, Gao, and Sang]{qing2024hierarchical}
Zhiwu Qing, Shiwei Zhang, Jiayu Wang, Xiang Wang, Yujie Wei, Yingya Zhang, Changxin Gao, and Nong Sang.
\newblock Hierarchical spatio-temporal decoupling for text-to-video generation.
\newblock In \emph{Proceedings of the IEEE/CVF Conference on Computer Vision and Pattern Recognition}, pages 6635--6645, 2024.

\bibitem[Rombach et~al.(2022)Rombach, Blattmann, Lorenz, Esser, and Ommer]{rombach2022high}
Robin Rombach, Andreas Blattmann, Dominik Lorenz, Patrick Esser, and Bj{\"o}rn Ommer.
\newblock High-resolution image synthesis with latent diffusion models.
\newblock In \emph{Proceedings of the IEEE/CVF conference on computer vision and pattern recognition}, pages 10684--10695, 2022.

\bibitem[Ronneberger et~al.(2015)Ronneberger, Fischer, and Brox]{ronneberger2015u}
Olaf Ronneberger, Philipp Fischer, and Thomas Brox.
\newblock U-net: Convolutional networks for biomedical image segmentation.
\newblock In \emph{Medical image computing and computer-assisted intervention--MICCAI 2015: 18th international conference, Munich, Germany, October 5-9, 2015, proceedings, part III 18}, pages 234--241. Springer, 2015.

\bibitem[Saito et~al.(2020)Saito, Saito, Koyama, and Kobayashi]{saito2020train}
Masaki Saito, Shunta Saito, Masanori Koyama, and Sosuke Kobayashi.
\newblock Train sparsely, generate densely: Memory-efficient unsupervised training of high-resolution temporal gan.
\newblock \emph{International Journal of Computer Vision}, 128\penalty0 (10):\penalty0 2586--2606, 2020.

\bibitem[Soomro et~al.(2012)Soomro, Zamir, and Shah]{soomro2012dataset}
Khurram Soomro, Amir~Roshan Zamir, and Mubarak Shah.
\newblock A dataset of 101 human action classes from videos in the wild.
\newblock \emph{Center for Research in Computer Vision}, 2\penalty0 (11):\penalty0 1--7, 2012.

\bibitem[Tran et~al.(2015)Tran, Bourdev, Fergus, Torresani, and Paluri]{tran2015learning}
Du Tran, Lubomir Bourdev, Rob Fergus, Lorenzo Torresani, and Manohar Paluri.
\newblock Learning spatiotemporal features with 3d convolutional networks.
\newblock In \emph{Proceedings of the IEEE international conference on computer vision}, pages 4489--4497, 2015.

\bibitem[Unterthiner et~al.(2018)Unterthiner, Van~Steenkiste, Kurach, Marinier, Michalski, and Gelly]{unterthiner2018towards}
Thomas Unterthiner, Sjoerd Van~Steenkiste, Karol Kurach, Raphael Marinier, Marcin Michalski, and Sylvain Gelly.
\newblock Towards accurate generative models of video: A new metric \& challenges.
\newblock \emph{arXiv preprint arXiv:1812.01717}, 2018.

\bibitem[Van Den~Oord et~al.(2017)Van Den~Oord, Vinyals, et~al.]{van2017neural}
Aaron Van Den~Oord, Oriol Vinyals, et~al.
\newblock Neural discrete representation learning.
\newblock \emph{Advances in neural information processing systems}, 30, 2017.

\bibitem[Wang et~al.()Wang, Ge, Chen, Zhou, Wang, Cheng, and Yuan]{Prompt2Poster}
Shaodong Wang, Yunyang Ge, Liuhan Chen, Haiyang Zhou, Qian Wang, Xinhua Cheng, and Li Yuan.
\newblock Prompt2poster: Automatically artistic chinese poster creation from prompt only.
\newblock In \emph{ACM Multimedia 2024}.

\bibitem[Wang et~al.(2024)Wang, Yuan, Zhang, Chen, Wang, Zhang, Shen, Zhao, and Zhou]{wang2024videocomposer}
Xiang Wang, Hangjie Yuan, Shiwei Zhang, Dayou Chen, Jiuniu Wang, Yingya Zhang, Yujun Shen, Deli Zhao, and Jingren Zhou.
\newblock Videocomposer: Compositional video synthesis with motion controllability.
\newblock \emph{Advances in Neural Information Processing Systems}, 36, 2024.

\bibitem[Wang et~al.(2004)Wang, Bovik, Sheikh, and Simoncelli]{wang2004image}
Zhou Wang, Alan~C Bovik, Hamid~R Sheikh, and Eero~P Simoncelli.
\newblock Image quality assessment: from error visibility to structural similarity.
\newblock \emph{IEEE transactions on image processing}, 13\penalty0 (4):\penalty0 600--612, 2004.

\bibitem[Xiong et~al.(2018)Xiong, Luo, Ma, Liu, and Luo]{xiong2018learning}
Wei Xiong, Wenhan Luo, Lin Ma, Wei Liu, and Jiebo Luo.
\newblock Learning to generate time-lapse videos using multi-stage dynamic generative adversarial networks.
\newblock In \emph{Proceedings of the IEEE Conference on Computer Vision and Pattern Recognition}, pages 2364--2373, 2018.

\bibitem[Yan et~al.(2021)Yan, Zhang, Abbeel, and Srinivas]{yan2021videogpt}
Wilson Yan, Yunzhi Zhang, Pieter Abbeel, and Aravind Srinivas.
\newblock Videogpt: Video generation using vq-vae and transformers.
\newblock \emph{arXiv preprint arXiv:2104.10157}, 2021.

\bibitem[Yu et~al.(2023)Yu, Cheng, Sohn, Lezama, Zhang, Chang, Hauptmann, Yang, Hao, Essa, et~al.]{yu2023magvit}
Lijun Yu, Yong Cheng, Kihyuk Sohn, Jos{\'e} Lezama, Han Zhang, Huiwen Chang, Alexander~G Hauptmann, Ming-Hsuan Yang, Yuan Hao, Irfan Essa, et~al.
\newblock Magvit: Masked generative video transformer.
\newblock In \emph{Proceedings of the IEEE/CVF Conference on Computer Vision and Pattern Recognition}, pages 10459--10469, 2023.

\bibitem[Yu et~al.(2024)Yu, Lezama, Gundavarapu, Versari, Sohn, Minnen, Cheng, Gupta, Gu, Hauptmann, et~al.]{yu2023language}
Lijun Yu, Jos{\'e} Lezama, Nitesh~B Gundavarapu, Luca Versari, Kihyuk Sohn, David Minnen, Yong Cheng, Agrim Gupta, Xiuye Gu, Alexander~G Hauptmann, et~al.
\newblock Language model beats diffusion--tokenizer is key to visual generation.
\newblock \emph{International Conference on Learning Representations}, 2024.

\bibitem[Yuan et~al.(2024{\natexlab{a}})Yuan, Huang, Shi, Xu, Zhu, Lin, Cheng, Yuan, and Luo]{magictime}
Shenghai Yuan, Jinfa Huang, Yujun Shi, Yongqi Xu, Ruijie Zhu, Bin Lin, Xinhua Cheng, Li Yuan, and Jiebo Luo.
\newblock Magictime: Time-lapse video generation models as metamorphic simulators.
\newblock \emph{arXiv preprint arXiv:2404.05014}, 2024{\natexlab{a}}.

\bibitem[Yuan et~al.(2024{\natexlab{b}})Yuan, Huang, Xu, Liu, Zhang, Shi, Zhu, Cheng, Luo, and Yuan]{ChronoMagic-Bench}
Shenghai Yuan, Jinfa Huang, Yongqi Xu, Yaoyang Liu, Shaofeng Zhang, Yujun Shi, Ruijie Zhu, Xinhua Cheng, Jiebo Luo, and Li Yuan.
\newblock Chronomagic-bench: A benchmark for metamorphic evaluation of text-to-time-lapse video generation.
\newblock \emph{arXiv preprint arXiv:2406.18522}, 2024{\natexlab{b}}.

\bibitem[Zhang et~al.(2018)Zhang, Isola, Efros, Shechtman, and Wang]{zhang2018unreasonable}
Richard Zhang, Phillip Isola, Alexei~A Efros, Eli Shechtman, and Oliver Wang.
\newblock The unreasonable effectiveness of deep features as a perceptual metric.
\newblock In \emph{Proceedings of the IEEE conference on computer vision and pattern recognition}, pages 586--595, 2018.

\bibitem[Zhao et~al.(2024)Zhao, Zhang, Cun, Yang, Niu, Li, Hu, and Shan]{zhao2024cv}
Sijie Zhao, Yong Zhang, Xiaodong Cun, Shaoshu Yang, Muyao Niu, Xiaoyu Li, Wenbo Hu, and Ying Shan.
\newblock Cv-vae: A compatible video vae for latent generative video models.
\newblock \emph{arXiv preprint arXiv:2405.20279}, 2024.

\bibitem[Zheng et~al.(2024)Zheng, Peng, Yang, Shen, Li, Liu, Zhou, Li, and You]{opensora}
Zangwei Zheng, Xiangyu Peng, Tianji Yang, Chenhui Shen, Shenggui Li, Hongxin Liu, Yukun Zhou, Tianyi Li, and Yang You.
\newblock Open-sora: Democratizing efficient video production for all, 2024.

\end{thebibliography}


\end{document}